\documentclass[letterpaper, 10 pt, conference]{ieeeconf}  

\IEEEoverridecommandlockouts                              

\usepackage{amsmath} 
\usepackage{amssymb}  
\usepackage{adjustbox}
\usepackage[utf8]{inputenc} 
\usepackage[T1]{fontenc}    
\usepackage{hyperref}       
\usepackage{url}            
\usepackage{booktabs}       
\usepackage{amsfonts}       
\usepackage{nicefrac}       
\usepackage{stfloats}
\usepackage{microtype}      
\usepackage{xcolor}         
\usepackage{float}
\usepackage{placeins}
\newcommand{\expect}[2]{\mathop{\mathbb{E}}_{#2} \left[ {#1} \right]}
\usepackage{algorithm}
\usepackage{algpseudocode}

\usepackage{caption}
\usepackage{subcaption}
\usepackage{multirow}
\usepackage{bbold}

\title{\LARGE \bf
Vehicle Type Specific Waypoint Generation
}

\author{Yunpeng Liu$^{1,2}$
Jonathan Wilder Lavington$^{1,2}$ 
Adam \'Scibior$^{1,2}$
Frank Wood$^{1,2,3}$
\thanks{$^{1}$Inverted AI,
$^{2}$University of British Columbia,
$^{3}$Mila}
}

\begin{document}

\maketitle
\thispagestyle{empty}
\pagestyle{empty}

\begin{abstract}
We develop a generic mechanism for generating vehicle-type specific sequences of waypoints from a probabilistic foundation model of driving behavior. Many foundation behavior models are trained on data that does not include vehicle information, which limits their utility in downstream applications such as planning. Our novel methodology conditionally specializes such a behavior predictive model to a vehicle-type by utilizing byproducts of the reinforcement learning algorithms used to produce vehicle specific controllers. We show how to compose a vehicle specific value function estimate with a generic probabilistic behavior model to generate vehicle-type specific waypoint sequences that are more likely to be physically plausible then their vehicle-agnostic counterparts. 
\end{abstract}

\section{INTRODUCTION} 

Behavior models \cite{9158529} are useful for planning \cite{sadat_jointly_2019} and control \cite{vo2009behavior}.  One use of such models is to do route planning in the presence of other stochastic agents whose types and goals may not be known~\cite{cui2021lookout}. Another is to, for instance, conditionally propose a sequence of waypoints to be followed by a controlled ego vehicle which can be used to evaluate the safety or performance of another egocentric controller~\cite{agarwal2019learning}.  In all cases the sequence of generated waypoints is generally handed to a lower-level policy that actuates the physical controls of the vehicle to ``achieve'' or ``follow'' these waypoints~\cite{chen_learning_2020}. In this work we address the discrepancy in performance which can be observed when a ``generic'' behavior model, trained to conditionally propose a sequence of waypoints to be followed for the purpose of model evaluation and testing, is applied to a vehicle-type specific lower-level controller.  

In this setting, our generic behavior model is defined with a probabilistic conditional model of multi-agent trajectories trained on data that does not include labelling of vehicle specific parameters~\cite{interactiondataset}.  Realistically such labeled data can't be collected in the real world as it would require having direct access to labeled measurements of vehicle parameters for every vehicle encountered. For this reason, behavior models are usually trained on overhead bird-view data~\cite{suo2021trafficsim} or ego-centric bounding box data~\cite{yao2019egocentric}. Here, only basic information of each vehicle is known, and nothing about the vehicles' masses, torque curves, or other parameters is directly observable. 

Such behavior models can be framed as multi-agent policies, and are defined as a mapping from a history of observations to an action that determines the next state. Typically in this setting, state includes all agent positions and orientations and a map in some representation. In some cases, such policies are even explicitly factorized as the product of individual agents interacting in a shared, deterministic world \cite{suo2021trafficsim,  9565113}. When the agents that comprise these models cannot directly observe aspects of true world state, they are referred to as partially observing agents. Generic foundation behavior models are partially observing in that they are often blind to vehicle specific dynamics, and as a result, must form an internal ``belief" over what the state of the system is, including the types of all vehicles, given the observations they can make. The internal uncertainty of such models can additionally stem from the multi-agent setting itself, where the policy must infer driver intent~\cite{zhao2020tnt}, style, and more~\cite{7995831}. 
Such partially observing behavior models become particularly problematic when they are used to provide waypoints for a lower-level controller of a particular vehicle-type to follow. Consider the following example: in self-driving car simulations, it is desirable for all non-ego agents to behave in highly realistic ways to avoid sim-to-real issues in training and validation \cite{chen2022understanding}. This means both that the vehicles must behave realistically with respect to physical behavior (accelerations, steering alignments, and wheel rotations all being physically realistic), while also being behaviorally realistic (with the patterns and distribution of behaviors matching human-like behaviors). A generic behavior model can generally produce realistically distributed behaviors in aggregate, but often times cannot produce physically realistic driving between those waypoints. To ensure its distribution over waypoints is physically tractable, such a model must be coupled to a lower-level controller that has been trained to follow sequences of waypoints through interaction with a realistic version of vehicle dynamics and road conditions~\cite{chen_learning_2020}.

The problem we address in this work arises at this intersection: we show how to condition a behavioral model so as to generate sequences of waypoints that are more likely to be achievable by specific vehicle types.  For instance, if we know that the vehicle to be controlled is a heavy truck with a weak engine, we would like to automatically produce behaviorally diverse and realistic waypoint sequences that are dynamically achievable by this heavy truck using a lower-level controller trained to drive such a vehicle. 

To achieve this, we first generate a dataset of achievable waypoints for several vehicle types by manually driving each vehicle in simulation.  This manual driving is done such that we explore a wide envelope of dynamics for each vehicle type. From these examples, we construct a ``waypoint following'' reinforcement learning environment in which we train a vehicle specific policy and value function that learns how to apply steering and acceleration controls. More specifically, this vehicle learns to take actions so as to match reference trajectories at all points in both space and time, given both the current state, and a sub-sequence of goal waypoints ahead. We then show how to use the learned value function for a particular vehicle type to condition a behavioral model to produce waypoint sequences that are achievable trajectories for the associated vehicle type. We integrate our work into the CARLA \cite{dosovitskiy2017carla} open source driving simulator.

\begin{figure*}[tp]
\vspace{1.5 mm}
     \centering
     \begin{subfigure}[b]{0.225\textwidth}
        \centering
        \includegraphics[width=1\linewidth,keepaspectratio]{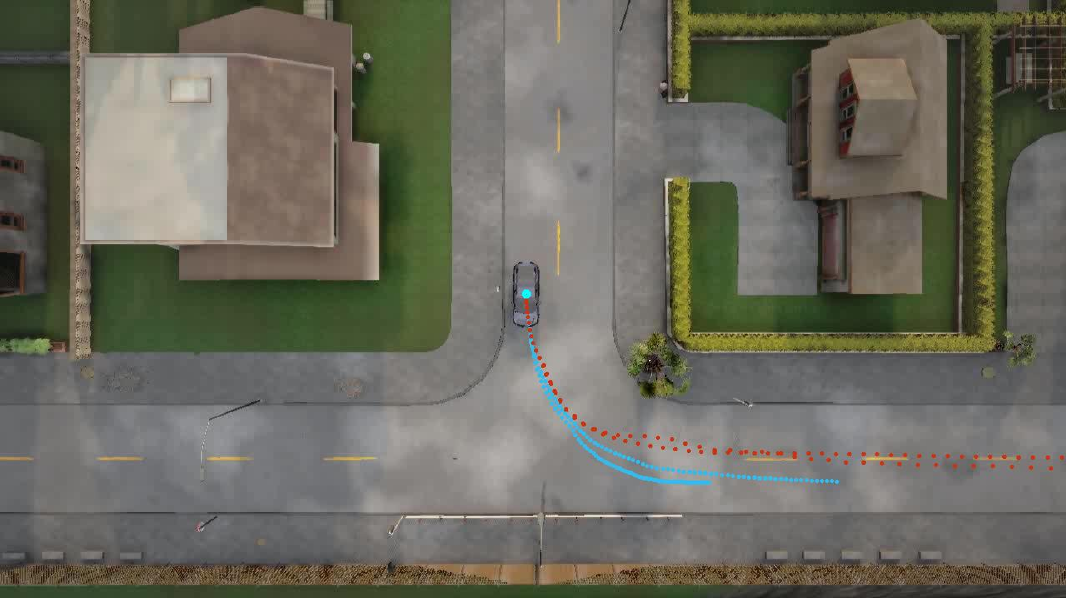}
        \label{fig:is_trajectories:1}
    \end{subfigure} 
    \begin{subfigure}[b]{0.225\textwidth}
        \centering
        \includegraphics[width=1\linewidth,keepaspectratio]{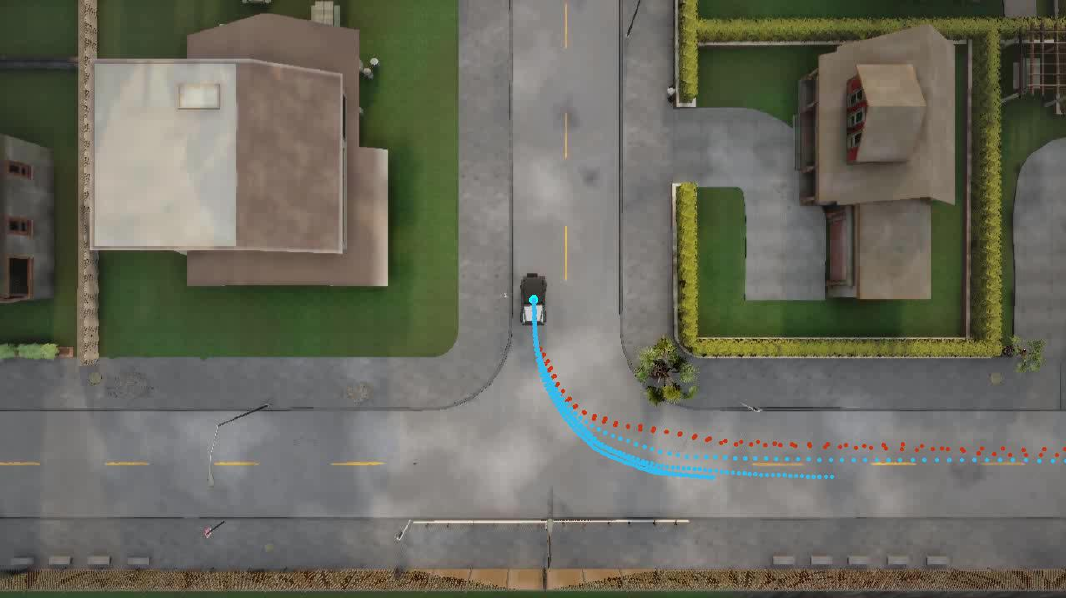}
         \label{fig:is_trajectories:2}
    \end{subfigure}
        \begin{subfigure}[b]{0.225\textwidth}
        \centering
        \includegraphics[width=1\linewidth,keepaspectratio]{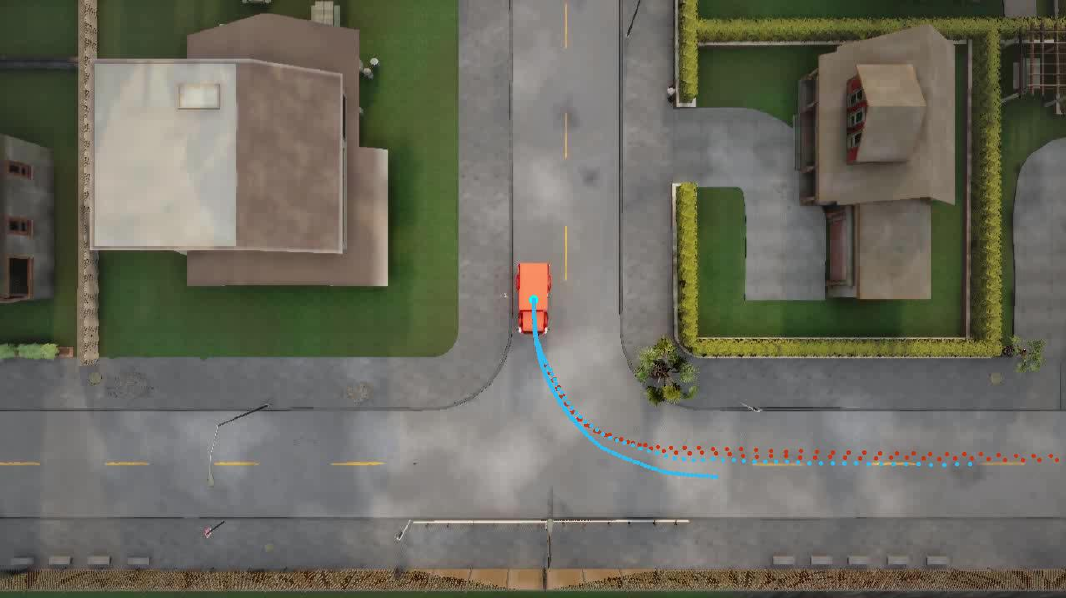}
         \label{fig:is_trajectories:3}
    \end{subfigure}
        \begin{subfigure}[b]{0.225\textwidth}
        \centering
        \includegraphics[width=1\linewidth,keepaspectratio]{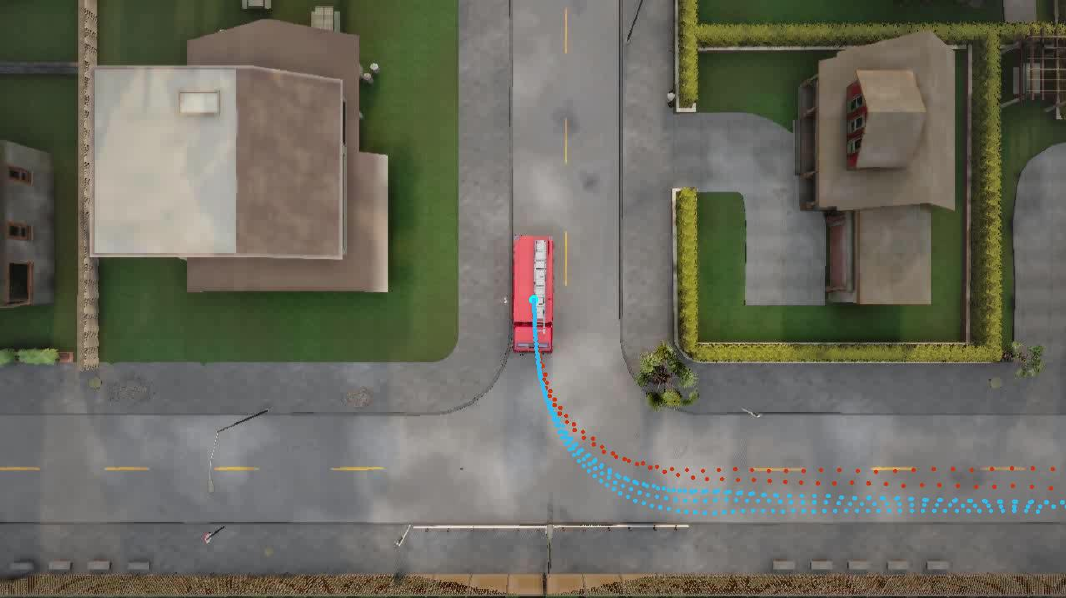}
         \label{fig:is_trajectories:4}
    \end{subfigure}
            \begin{subfigure}[b]{0.225\textwidth}
        \centering
        \includegraphics[width=1\linewidth,keepaspectratio]{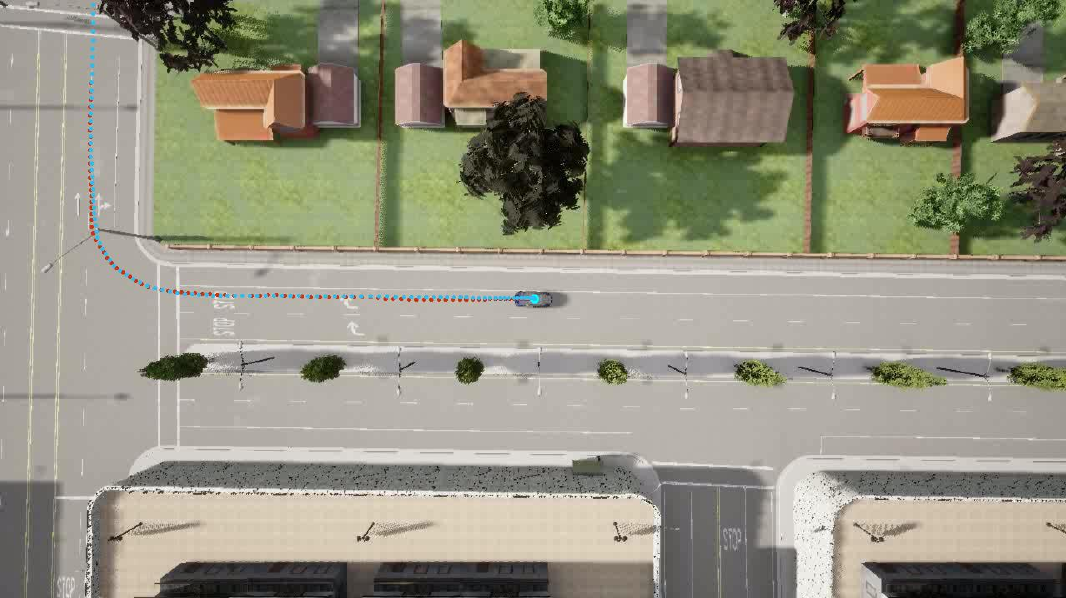}
         \label{fig:ref_trajectories:1}
    \end{subfigure}
                \begin{subfigure}[b]{0.225\textwidth}
        \centering
        \includegraphics[width=1\linewidth,keepaspectratio]{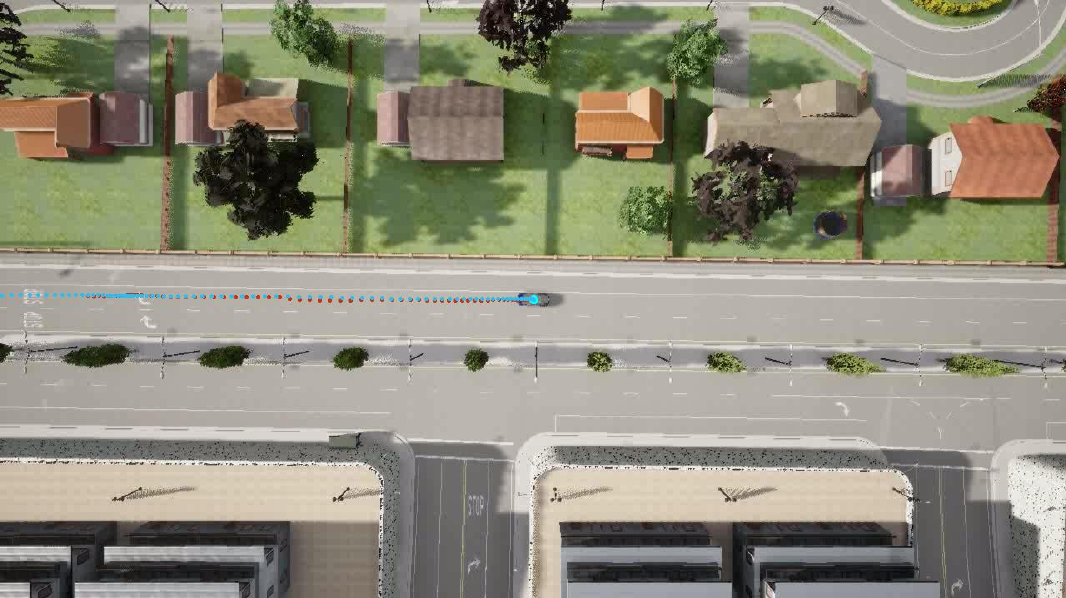}
         \label{fig:ref_trajectories:2}
    \end{subfigure}
                \begin{subfigure}[b]{0.225\textwidth}
        \centering
        \includegraphics[width=1\linewidth,keepaspectratio]{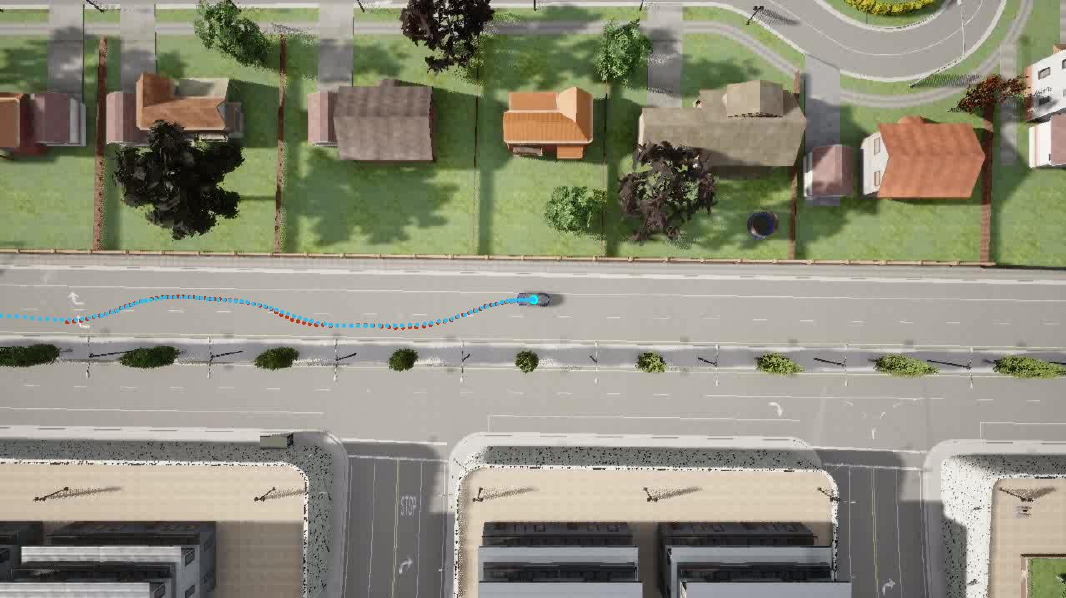}
         \label{fig:ref_trajectories:3}
    \end{subfigure}
                \begin{subfigure}[b]{0.225\textwidth}
        \centering
        \includegraphics[width=1\linewidth,keepaspectratio]{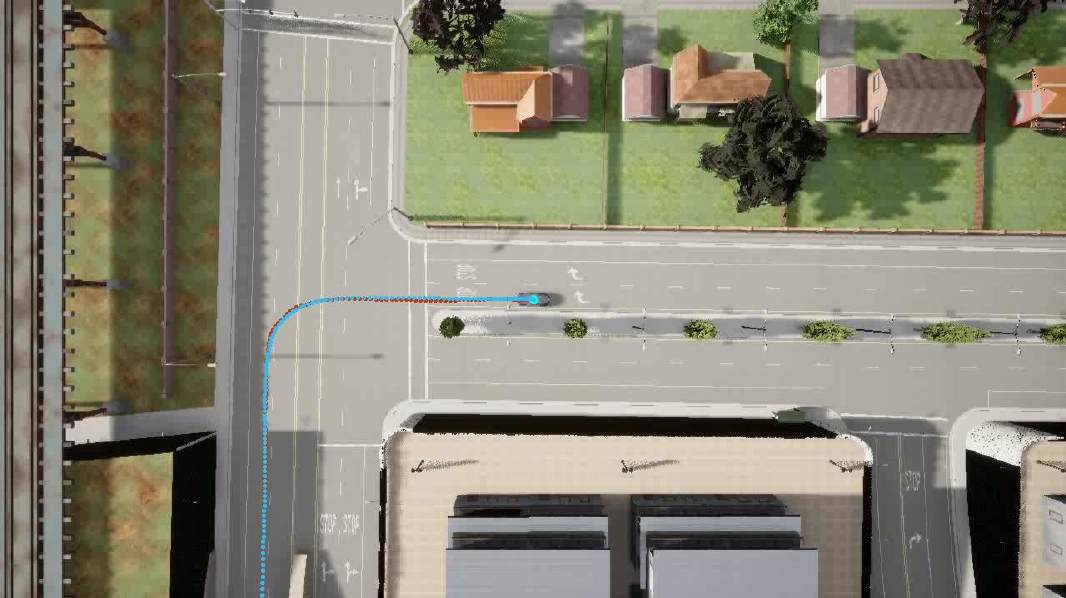}
         \label{fig:ref_trajectories:4}
    \end{subfigure}
    
    \caption{\textbf{Top row} Importance sampled, feasible trajectories are marked in blue for the waypoint following agent of four different vehicle types (Tesla, Jeep, shipping truck, fire truck) in the CARLA simulator. Trajectories marked in red represent trajectories with weights that are close to zero, and thus are unlikely to be feasible.  \textbf{Bottom Row} Manually collected expert trajectories marked in blue to cover the agent's full action space inside the waypoint following environment. RL agent rollouts for each scenario are marked in red. Scenarios are similar across vehicle types. }
    \label{fig:traj-viz}
\end{figure*}

\section{METHODS}
\label{sec:methods}
Our methodology couples solutions to three sub-problems: defining an environment for training low-level vehicle controllers, using RL to train low-level waypoint following agents within this environment, and using artifacts from those training processes within a Bayesian procedure for conditioning a foundation behavioral model to produce vehicle specific waypoints. We start by defining the waypoint following environment in which we train our vehicle-type specific, low-level controllers. We then describe the reinforcement learning (RL) approach we use, in particular defining the reward, the Markov decision process (MDP), the value function, and the choice of RL algorithm.  We then review the approximate inference perspective on RL and explain our final, novel step. In this step we combine a vehicle-type specific value function with a foundation behavior model to form an approximate posterior distribution over trajectories that are achievable by a particular vehicle-type and its associated controller.

\subsection{Waypoint Following Environment}
In this paper we define a target trajectory as sequence of waypoints $k_{1:T}$, where each waypoint is a tuple of target coordinates, speed, and orientation $k_t = (x_{t}^{k}, y_{t}^{k}, v_{t}^{k}, \psi_{t}^{k})$ for $t \in 1,\dots,T$. The waypoint following task is to produce a sequence of actions $a_t, t=1,\ldots,T$, consisting of vehicle controls. These controls consist of steering, throttle, and brake, and must produce a sequence of realized vehicle positions $\{x^s_t, y^s_t\}_{t=0}^T$ which resemble the target trajectory as closely as possible. Our notion of success in following a target trajectory is that the vehicle stays within a distance $\epsilon$ away from its reference trajectory at all time-steps t. That is, we say that a vehicle successfully followed a target trajectory if for all $t$,
\vspace{-0.1cm}
\begin {align}
d_t := \sqrt{(x_{t}^{s} \!\!- \!\!x_{t}^{k})^2 + (y_{t}^{s} \!\!- \!\!y_{t}^{k})^2 } \leq \epsilon, \label{eq:success}
\end {align}
\vspace*{-0.1cm}
where the superscript $s$ indicates the positions generated through interaction with the simulator, and $k$ defines the desired positions in the reference trajectory. If there exists a sequence of actions that results in a vehicle successfully following a  trajectory from its initial state, we say that this trajectory is feasible for the particular vehicle from that specific initial state. A waypoint following environment consists of one or more waypoint following tasks which we also refer to as scenarios. We construct waypoint following environments for each vehicle type we consider where target trajectories are generated by manual driving. These manually driven trajectories are picked to maximize state-space coverage to ensure proper generalization of the policy.

Certain target trajectories may not be feasible due to the dynamics and kinematic characteristics of a particular vehicle type. For example, maximum torque or minimum turning radius vary between most vehicle-types and these among other things heavily influence feasibility of trajectories. Which trajectories are feasible therefore depends on which vehicle is trying to follow them. We assume a setting where we have access to a generic probabilistic model of target trajectories learned by observing the motion of multiple vehicle-types, and our goal is to produce a model of target trajectories that can be followed by a specific vehicle-type $e$. In order to do that, we first describe the RL algorithm and MDP used to learn a vehicle-specific waypoint following policy. We then show that the artifacts obtained in this process can be used to construct vehicle-specific target trajectory models without further interaction with the simulator.

\subsection{Reinforcement Learning for Waypoint Following}

First we construct a waypoint following policy $\pi(a_t | s_t)$, which creates controls based on the vehicle's state and the target trajectory. In order to use RL to construct a waypoint following policy, we need to frame the waypoint following problem as an MDP. Here actions are defined to be the vehicle controls, i.e.~steering angle and single combined pedal acceleration/braking control (where negative values indicate braking and positive values indicate acceleration). The reward is then defined as the negated distance to the target trajectory
\begin {align}
\!\!r_t =  \! \!  \left( \epsilon \!- \! \sqrt{(x_{t}^{s} \!\!- \!\!x_{t}^{k})^2 + (y_{t}^{s} \!\!- \!\!y_{t}^{k})^2 } \right) \label{eq:reward}.
\end {align}
The full MDP state contains the vehicle's absolute position, orientation, velocity, the target trajectory, and any internal vehicle state, in particular that of the engine, which may influence its future dynamics. To simplify the state representation, and ensure that the state can be applied to arbitrary vehicle types, we consider six primary features: the longitudinal and lateral velocity of the vehicle, its gear, and the relative longitudinal and lateral position and orientation of the next target waypoint in the reference frame attached to the vehicle. Additionally, to allow the policy to better anticipate future position and heading requirements, we additionally provide it with a window of $H$ target waypoints into the future. We denote this state-space of size $3(H+1)$ as
\begin{align}
    s_t &= \left[v_{lon,t}^s, v_{lat, t}^s, g_t, \Delta x^s_{t:t+H}, \Delta y^s_{t:t+H},  \Delta \psi^s_{t:t+H}  \right], \\ 
    & \!\!\!\!\!\!\!\! \begin{bmatrix}
           \Delta x^s_{t+j} \\
           \Delta y^s_{t+j} \\ 
    \end{bmatrix}
    = \begin{bmatrix}
           (x^k_{t+j+1} - x^s_t) \\
           (y^k_{t+j+1} - y^s_t) \\ 
      \end{bmatrix}^\top \!\!
      \begin{bmatrix}
           cos(\psi^s_t) \ -sin(\psi^s_t) \\
        sin(\psi^s_t) \ \ cos(\psi^s_t) \\ 
      \end{bmatrix},
\end{align}
for each time index displacement $j<H$, with $\Delta \psi^s_{t+j} = (\psi^k_{t+j+1} - \psi^s_t)$. We also assume a fixed distribution over initial states, defined by the initial velocity, position, orientation, and gear of the vehicle. Finally, we assume that the transition dynamics are defined by a deterministic external driving simulator (e.g.~CARLA~\cite{dosovitskiy2017carla} or NVIDIA Drive Sim~\cite{nvidia_developer_2021}). For convenience we also define $\tau_t := \{a_t,s_t,r_t\}$, $\tau = \tau_{1:T}$, and the distribution over fixed time-horizon trajectories
\begin{align}
    q_{\theta}(\tau) = p(s_0) \prod\nolimits_{t=0}^{T}p(s_{t+1}|s_t,a_t)\pi_{\theta}(a_t|s_t),\label{equ:background:mdp}
\end{align}
for a particular parametric policy $\pi_{\theta}$. In this paper we incorporate an additional termination condition defined by Equation \ref{eq:success}, as well as a maximum time-limit for interaction. 

For a fixed horizon MDP, the optimal policy is one which maximizes the expected reward ahead. When conditioned on state, this expectation defines the value function or critic
\begin{align}
&V^{\pi_{\theta}}(s_t) = \expect{\sum_{t'=t}^{T} r_{t'}(a_t',s_t') }{\tau_{t:T}\sim q_{\pi_{\theta}}(\tau_{t:T}|s_t) } .
\end{align}
As is typical in RL, we learn a parametric approximation to this value function using a neural network $V^{\phi}(s_t)$ with parameter $\phi$, which takes the same inputs as the policy. Generally the value function is trained jointly with the policy, and in this work we specifically use Proximal Policy Optimization \cite{schulman2017ppo}. Training separately for each vehicle type $e$, we obtain the approximately optimal waypoint following policy $\pi_{\theta_e}$ and the associated value function $V^{\phi_e}$. Figure \ref{fig:avg_three} shows that policies obtained in this fashion successfully follow target trajectories known to be feasible.

\subsection{Reinforcement Learning as Inference} \label{sec:rlai}
 
We will briefly discuss the RL as inference (RLAI) framework~\cite{levine2018reinforcement,lavington2021a} and define the mathematical tools used in later subsections. In this setting, RL is posed as an approximate posterior inference problem over the set of trajectories conditioned on how well the agent should perform. This creates an inference problem defined by a set of latent random variables (the trajectories), and a set of observed random variables describing performance which are conditionally dependent on the state-action pair at time t. These observed random variables are called ``optimality" variables $O_t$, and are assumed to be Bernoulli distributed. When $O_t=1$ we say that the associated state-action was optimal, while if $O_t=0$ we say the state-action was sub-optimal. The parameters of these Bernoulli distributions are
\vspace*{-0.35cm}
\begin{align} p(O_t=1|a_t,s_t) &= \frac{1}{Z_0}\exp{r(a_t,s_t)},
\end{align}
\vspace*{-0.35cm}

where the normalizing constant $Z_0$ is dependent on the range of values the reward can take on. 
With optimality defined, we can consider the probability that a given trajectory, the sequence of state action pairs sampled from a policy interacting with the environment $\tau$, is itself optimal. For finite horizon problems, this joint distribution over both latent and observed random variables, $p(\tau,O_{1:T-1})$, is defined
\vspace{-0.225cm}
\begin{align}
    & p(s_1)\prod_{t=1}^{T-1} \Big[ p(s_{t+1}|s_{t},a_t)\pi_0(a_{t}|s_{t})p(O_t|\tau_t)  \Big],
\end{align}
\vspace*{-0.335cm}

where $\pi_0(a_t|s_t)$ denotes the `default' policy or prior over actions given states. In general, using this joint distribution to derive an exact posterior is intractable, therefore we rely on variational inference or Monte-Carlo methods.

\subsection{Evaluating Feasibility}

We are interested in the distribution of trajectories reachable by a particular vehicle type $e$.  We define a posterior over feasible trajectories, conditioned on optimality, initial conditions, and the vehicle type $e$. Here the optimality variables depend on how well the agent was able to reproduce the reference policy through the definition of the reward. The initial conditions $c = (c_{-U}, \ldots, c_0)$ consist of a `burn-in' sequence of states of length $U+1$ where each element is similar in nature to part of $s_t$ combined with a waypoint $k_t^{j}$ from some empirical distribution over burn-in waypoints
\vspace*{-0.2cm}
\begin{equation}
    c_t = [v_{lon,t}^s, v_{lat, t}^s, g_t,x_{t}^{j}, y_{t}^{j}, v_{t}^{j}, \psi_{t}^{j}].
\end{equation}
\vspace*{-0.2cm}
Which means the posterior is defined according to
\begin{equation}
    p_e(k_{1:T}|O_{1:T},c) = \frac{p_e(O_{1:T}|k_{1:T}, c) p(k_{1:T}|c)}{p_e(O_{1:T}|c)}.
\end{equation} 
\vspace*{-0.4cm}

In order to characterize this distribution, we start with an estimate of the likelihood term $p_e(O_{1:T} | c, k_{1:T})$, which we arrive at via the the probabilistic value function~\cite{piche2018probabilistic,lavington2021a}
\begin {align*}
 &p_e(\textsl{O}_{1:T} | k_{1:T}, c) = \int p_e(O_{1:T}, \tau | k_{1:T}, c) d{\tau}  \\
 &= \int \prod_{t=0}^{T-1} \frac{\exp \left(r(s_t, a_t)\right) }{Z_0}\pi_{\theta_{e}}(a_t | s_t )p(s_{t+1} | s_{t}, a_{t}) \mathbb{1}_{[d_t\leq\epsilon]}  d{\tau} \\
 &= \frac{1}{Z} \expect{\exp\left[\sum^{T-1}_{t=0} r_{t}(a_t, s_t) \right]}{\tau|s_0 \sim q_{\pi_{\theta_e}}},
\end{align*}
where $Z=\prod_{t=0}^{T-1} Z_0$. As a reminder, note that the reference trajectory $k_{1:T}$ is consumed by both $r(s_t,a_t)$ and $d_t$. Here we include an indicator function $\mathbb{1}_{[d_t\leq\epsilon]}$ to denote that the posterior only includes support over trajectories which are feasible. We then take the log of both sides, to give a expectation that is more familiar
\begin{align}
    & \log p_e(\textsl{O}_{1:T} | k_{1:T}, c) = \log p_e(O_{1:T} | s_0) \nonumber \\
     &=  \log\left[  \expect{\exp\left[\sum^{T-1}_{t=0} r_{t}(a_t, s_t) \right]}{\tau|s_0 \sim q_{\pi_{\theta_e}}}\right]-\log(Z). 
\end{align}
Using Jensen's inequality we recover an upper bound to the artifact that classical RL algorithms train
\begin{align}
\log p_e(O_{1:T} | s_0) &\geq \expect{\sum^{T-1}_{t=0} r_{t}}{\tau|s_0 \sim q_{\pi_{\theta_e}}}-\log(Z) \label{eq:bound} \\
&= V^{\pi_{\theta_e}}(s_0)  -\log(Z)  
\end{align}  
Using this value function, we produce a surrogate likelihood function along with its associated normalizing constant,
\begin{align}
 p_e(O_{1:T}|c) &= \mathop{\mathbb{E}}_{p_e(k_{1:T}|c)} \left[ \exp(V^{\pi_{\theta_e}}(s_0) \right]-\log(Z).
\end{align} 
Because both $p(O_{1:T} |c) $ and $p(O_{1:T} | s_0)$ are now easy to estimate and $p(s_0|c)$ can be sampled from directly, we use importance sampling to gather examples which are close to the desired posterior distribution
\begin{equation}
   p_e(k_{1:T}|O_{1:T},c) \approx  \frac{\exp(V^{\pi_{\theta_e}}(s_0))p_e(s_0|c)}{\mathop{\mathbb{E}}_{p_e(k_{1:T}|c)} \left[ \exp(V^{\pi_{\theta_e}}(s_0) \right]}.
   \label{eq:importance_w}
\end{equation}
Using the right hand side of \eqref{eq:importance_w} to score, we apply importance weighting on top of the vehicle planner proposals to increase the percentage of feasible trajectories generated for a particular vehicle type, as shown in Algorithm~\ref{alg:RPI}.
\begin{algorithm}[tp]
    \caption{Planner Refinement Via Re-sampling}
    \label{alg:RPI}
    \begin{algorithmic}[1]
        \State \textbf{Input:} Planner $p(k_{1:T})$, Value function $V_{\pi_{\theta_e}}(s_t)$, Number of Importance Samples $L$ 
        \State  \textbf{for} $\ell=0,1,2 ...$ do
            \State \medspace \medspace \medspace \medspace Sample rollout $k_{1:T}^{\ell}$ from $p(k_{1:T}| c)$
            \State  \medspace \medspace \medspace \medspace Evaluate un-normalized $w$  using equation \ref{eq:importance_w}, store.
        \State \textbf{end for}
        \State Select a rollout $k_{1:T}^{i^{\ell}}$ by sampling an index with probability proportional to the weights 
        \begin{align} 
        \resizebox{.35\textwidth}{!}{%
             $\!\!\!\!\!\!\!\!i^l \!\!\sim \text{Discrete} \left(\frac{w^1}{\sum_{\ell=1}^Lw^l}, \frac{w^2}{\sum_{\ell=1}^Lw^\ell}, \cdots, \!\frac{w^L}{\sum_{\ell=1}^Lw^\ell}  \right)$%
             }
             \label{eq: discrete weights}
        \end{align}  
        \State \textbf{Output:} Deploy agent using ITRA trajectory $k_{1:T}^{i^{\ell}}$.
    \end{algorithmic}
\end{algorithm}

\section{EXPERIMENTS}
\subsection{Performance of waypoint follower}
To obtain the value function for each vehicle type, we train four waypoint following agents which include: a fire truck, a small shipping truck, a Tesla Model 3 and a Jeep Wrangler. Each of these vehicles is provided by CARLA \cite{dosovitskiy2017carla}, a high fidelity simulator for autonomous driving. It provides a variety of vehicle blueprints with different dynamics. To train a vehicle specific waypoint following agent, we collect vehicle specific waypoints inside CARLA by driving manually as shown in Figure~\ref{fig:traj-viz}. We collect four different scenarios which were designed to cover the agent’s full action space for each vehicle type. The scenarios include left-turn, right-turn, fully-stopping and an S-shaped trajectory.  We record the following attributes for each vehicle type: position, yaw, velocity and action (steering angle and throttle) at each frame given by the simulator. Each record consists of 120 to 180 waypoints.

We train each vehicle to match manually gathered trajectories using proximal policy optimization (PPO)~\cite{schulman2017ppo}. As stated in section \ref{sec:methods}, we include in the state-space $H=30$ future goals for both training and evaluation. These future goals are pre-processed by a multilayer perceptron(MLP) layer before passing into the value function and policy network to stabilize the training. Figure \ref{fig:avg_three} shows the average percentage of waypoints hit for the 4 scenarios used in training by the RL agents of the 4 vehicles.  We note that the fire truck waypoint following agent only solves 3 out of 4 scenarios. We assume this follows from the fire truck being the least controllable vehicle, due to its longer braking distance, larger turning radius and slower acceleration compared to the other vehicle types. We note also that the accuracy of predicted values, given in the value loss in Figure \ref{fig:value_loss}, often provides a clear indication of how well that value function will score the feasibility of proposed trajectories.

\subsection{Performance of planner refinement}

\subsubsection{Datasets}
For each vehicle type, we record four different initial trajectories generated by executing CARLA autopilot over a fixed time interval, from a fixed location. Using these trajectories, we warm-start our behavioral model by executing the waypoint-following policy to the initial condition of interest, described in \ref{sec:methods}. From this position, and using the previous data along that trajectory, we then produce an observed sequence of positions and velocities using our behavioral model. The examples are generated in a non-trivial setting, where the agent must complete exiting a roundabout or turn at an intersection. This experiment is executed for four auto-pilot trajectories and applied to four vehicles. We generate nine second trajectories given one second of observations at ten frames per second from ITRA following the kinematic parameter and evaluate what percentage of this route can be completed by the pretrained controller.   

\subsubsection{Evaluation Details}

\begin{figure*}[t]
     \centering
     \begin{subfigure}[c]{0.32\textwidth}

        \includegraphics[width=1\linewidth,keepaspectratio]{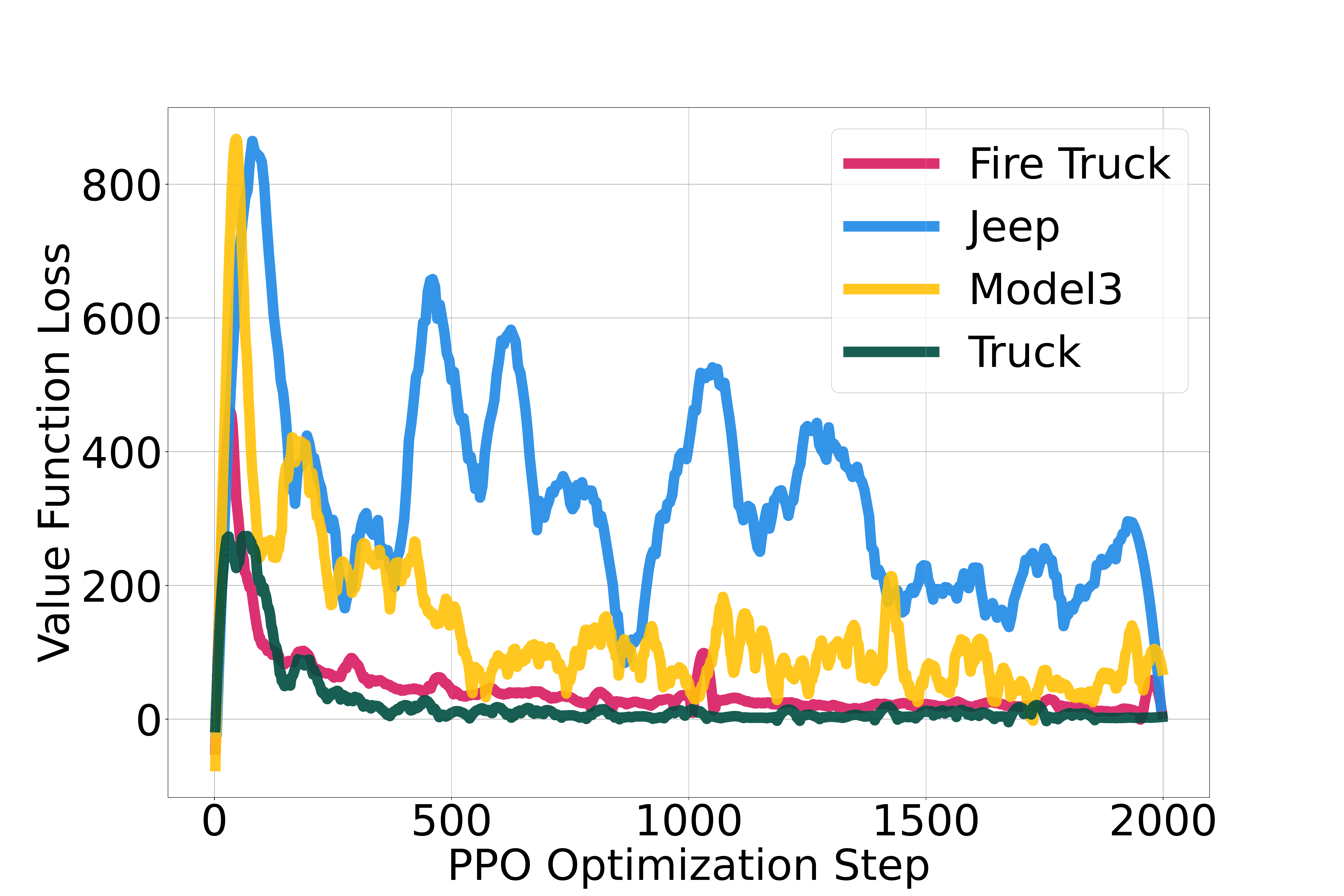}
        \caption{}
        \label{fig:value_loss}
    \end{subfigure} 
    \begin{subfigure}[c]{0.32\textwidth}

        \includegraphics[width=1\linewidth,keepaspectratio]{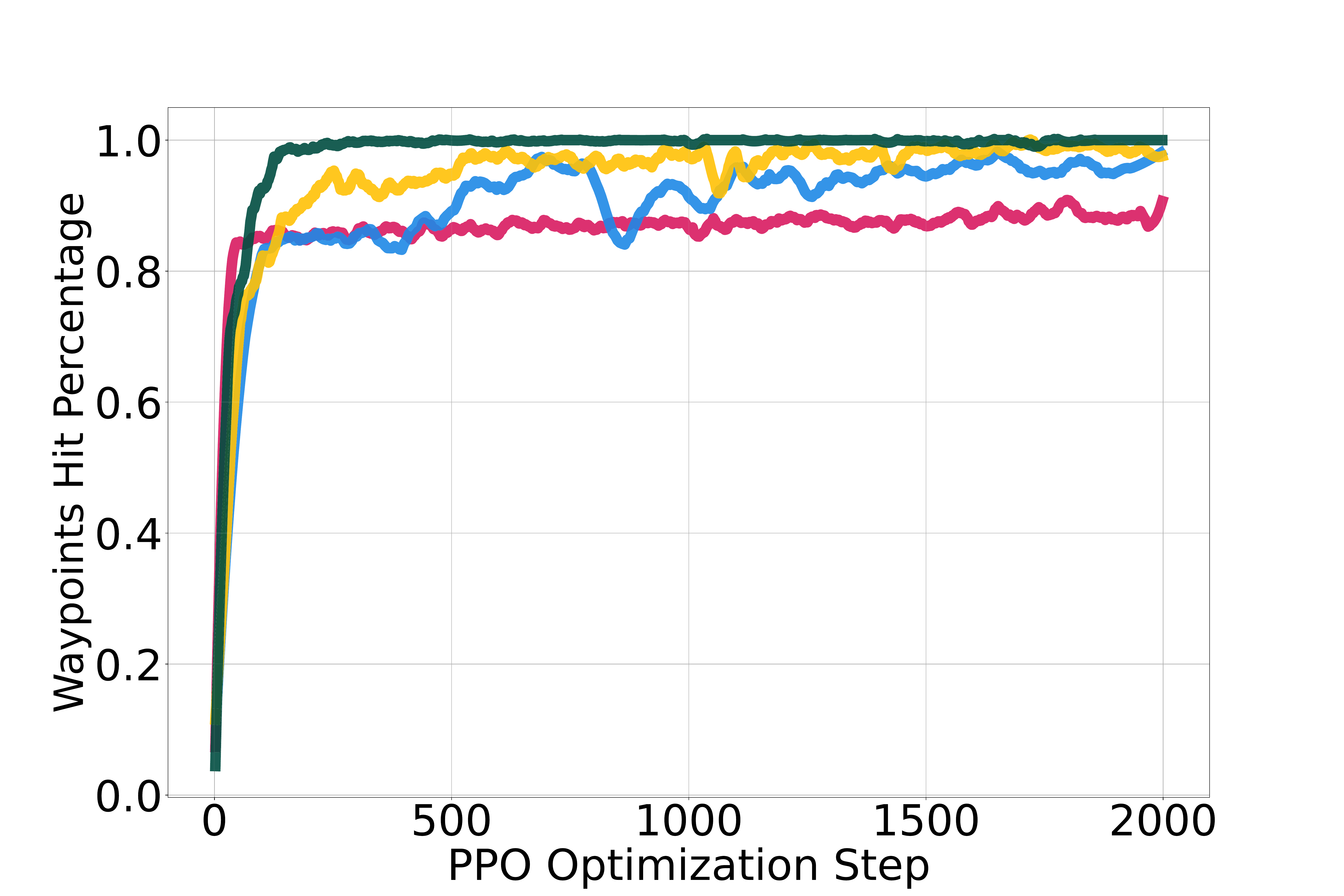}
        \caption{}
        \label{fig:avg_three}
        
    \end{subfigure}
    \begin{subfigure}[t]{0.345\textwidth}
    \resizebox{\columnwidth}{!}{%
        \begin{tabular}{||c | c |  c c c c c||} 
     \hline
     Vehicle Type & $\epsilon$&Distribution & IC 1 & IC 2 & IC 3 & IC 4 \\ [0.5ex]
     \hline\hline
     \multirow{0}{*}{Truck} &1.0 &Prior  &0.9 & 1.0 & 0.8& 0.9\\ 
     \hline
     &1.0&Posterior & \textbf{1.0} & 1.0& \textbf{1.0}& \textbf{1.0}\\
     \hline
     \hline
     \multirow{0}{*}{Model3}&1.0 &Prior  & 1.0 & 1.0 & 0.9 & 0.9 \\ 
     \hline
     &1.0&Posterior & 1.0 & 1.0 & \textbf{1.0} & \textbf{1.0}\\
     \hline
     \hline
      \multirow{0}{*}{Jeep} &1.0&Prior  & 1.0& 1.0 & 0.8& 1.0\\ 
     \hline
     &1.0&Posterior& 1.0 & 1.0& \textbf{1.0}& 1.0\\
      \hline
       \hline
      \multirow{0}{*}{Fire Truck}&1.5 &Prior  & 0.7 & 0.7 & 0.6 & 0.6 \\ 
     \hline
     &1.5&Posterior & \textbf{0.9} &\textbf{0.8} & \textbf{0.7} & \textbf{0.8}\\
     \hline
    \end{tabular}%
    }
     \caption{}
    \label{tab:refinement_result}
    \end{subfigure}
    \caption{Figure \ref{fig:value_loss} displays the value function MSE loss for 4 different vehicle types.  The value loss for the Jeep waypoint following agent fluctuates more than the other agents but eventually converges as it is sensitive to control inputs. Figure \ref{fig:avg_three} displays average percentage of waypoints solved across 4 different scenarios for all vehicle specific waypoint following agents. The waypoint following agent for the fire truck reaches all waypoints for 3 of 4 training scenarios but only reaches half of the waypoints for the sharp-braking scenario. Table \ref{tab:refinement_result} shows mean percentage of waypoints hit for the posterior $p_e(k_{1:T}| O_{1:T},c)$ compared to the prior $p(k_{1:T}|c)$ for different vehicle types $e$, and end conditions defined by Equation \ref{eq:success} given the distance $\epsilon$. Note that the Fire Truck end conditions are loosened to improve the interpretability of our results. Initial conditions(IC) 1 and 2 are exiting a roundabout and IC 3 and 4 are completing a turn. The Truck is able to follow posterior trajectories versus trajectories generated by the underlying behavior model prior across all initial conditions. The Fire Truck is able to follow more waypoints of the posterior than the prior even though its RL agent only solves 3 out of 4 training scenarios. We find that the Jeep and Tesla agents were already able to reach almost all waypoints generated from the prior for IC 1 and 2, thus the posterior only marginally improves the percentage of waypoints hit with a difference below significant digit level, though for IC 3 and 4, we observe the performance difference between the posteriors and prior.}
\end{figure*}

Following the initial burn in period described above, we sample 100 trajectories from the prior behavioral model ITRA for each vehicle type at 10Hz. We then execute the waypoint follower on these trajectories to get the percentage of the route completed for each of vehicle type. Rejection sampling is applied to the prior to avoid any off-road examples at evaluation time. The average percentage of waypoints hit for this baseline is shown this ``prior" row of Table \ref{tab:refinement_result}. To evaluate how effectively the value function would have avoided infeasible routes, we calculate the importance weighted expected value given in Table \ref{tab:refinement_result}.

\section{Related Work}
Behavioral models predict future motion of vehicles at different levels of abstraction, ranging from discrete, high-level commands \cite{ajanovic2018search} to specific vehicle positions at each time step \cite{  Rhinehart2020Deep, suo2021trafficsim,  zhao_tnt_nodate}. As these models are typically learned from external observations of vehicles with unknown attributes, they tend to disregard an explicit notion of feasibility, though some models \cite{9565113, suo2021trafficsim} incorporate constraints~\cite{7995816}. As shown in this work, simple constraints are not sufficient to guarantee feasibility of predicted trajectories for varying vehicle types. This is typically not a concern, since such models are usually deployed to predict the behavior of other agents in order to facilitate planning \cite{rudenko_human_2020}, but it becomes an issue when a behavioral model is used as a planner.

Vehicles operating in the real world typically feature a specialized planner which needs to be extremely reliable, so learned behavior models are generally not suitable for this purpose. Nonetheless, when generating high-fidelity simulation, it makes sense to use behavioral models as planners for non-playable characters. However, for predicted trajectories to be useful as plans, they must be feasible. Incorporating explicit constraints into trajectory planning has been extensively studied in the literature~\cite{jiang_efcient_nodate,wang_learning-based_nodate} but it typically makes assumptions on the vehicle dynamics, and relies on knowing problem specific constants. In our approach the constraints, and the dynamics which produce them, are not explicitly known. Instead, we estimate them by observing which trajectories a trained controller is able to follow.

Trajectory planning can be cast as a constrained optimization problem~\cite{sadat_jointly_2019}, where the target trajectory which satisfies a known dynamics model is selected to optimize a cost function. This cost function defined to produce desired characteristics, such as feasibility, comfort, achieving target, or minimizing infractions. Our framework, which uses maximum aposteriori estimates is equivalent to using a soft constraint instead of a direct projection onto the constraint set. While inexact, our method can easily maintain a set of diverse trajectories, which is crucial in the generation of diverse sets of scenarios.

Value functions learned via inverse RL have been used to produce rule-based planners with Bayesian optimization \cite{jiang_efcient_nodate}. By contrast, our approach only requires solving a single RL problem per vehicle, instead of a sequence of RL problems as is the case in most IRL methods. Unlike \cite{jiang_efcient_nodate}, we target feasibility of the trajectory, rather than human preferences.

Using learned behavioral models to create realistic simulations is a relatively recent research area \cite{suo2021trafficsim, 9565113}. Existing work uses simple 2D simulators with easily solvable kinematic models or performs 3D simulation by teleporting non-playable characters to their target positions at each time step. To the best of our knowledge, this is the first paper studying the task of incorporating realistic vehicle dynamics into simulations generated with learned behavioral models. Notably, hand-crafted behavioral models with corresponding controllers have been used for this purpose before~\cite{chao2020survey, kesting2009agents}. This includes controllers, but not behavioral models, which were trained using recorded human trajectories~\cite{9580542}.

\section{DISCUSSION}
In this work, we identify and formalize a fundamental problem at the intersection of foundational models of behavior and low level control. This issue, described by the discrepancy between a behavioral model and the physical constraints of a real (or simulated) environment, can be solved using artifacts already available from the byproducts of the reinforcement learning algorithm used to train low-level vehicle controllers. Our work illustrates that these byproducts implicitly target the same likelihoods needed to score feasibility, and in practice can be used to great effect. 

There remain however, opportunities to improve upon our work.  These include investigating methods to improve the bound in Equation~\ref{eq:bound}, amortized characterization of the posterior distribution of reachable waypoint trajectories given by Equation~\ref{eq:importance_w}, and meta-learning a {\em single} conditional distribution $p(k_{1_T}|O_{1:T},c,e)$ that automatically adapts to the given vehicle type.  We expect the approach we have proposed will generalize to additional vehicle types but this could also be verified by further study.

\section{ACKNOWLEDGEMENT}
We acknowledge the support of the Natural Sciences and Engineering Research Council of Canada (NSERC), the Canada CIFAR AI Chairs Program, and the Intel Parallel Computing Centers program. Additional support was provided by UBC's Composites Research Network (CRN), and Data Science Institute (DSI). This research was enabled in part by technical support and computational resources provided by WestGrid (www.westgrid.ca), Compute Canada (www.computecanada.ca), and Advanced Research Computing at the University of British Columbia (arc.ubc.ca).

\bibliography{main.bib, additional_zetora, iai-refs}
\bibliographystyle{unsrt}

\addtolength{\textheight}{-12cm}   



\end{document}